\documentclass[letterpaper, 10 pt, conference, nofonttune]{ieeeconf}  

\IEEEoverridecommandlockouts                              

\usepackage{cite}
\usepackage{amsmath,amssymb,amsfonts}
\usepackage{algorithm}
\usepackage{algpseudocode}
\usepackage{graphicx}
\usepackage{textcomp}
\usepackage{color}
\usepackage{bm}
\usepackage{comment}
\usepackage{makecell}
\usepackage{footnote}
\usepackage{subfigure}
\usepackage{balance}
\usepackage{microtype}

\usepackage[top=17.6mm, left=16.2mm, right=16.2mm, bottom=25.8mm]{geometry}

\title{\LARGE \bf
Learning Robot Inverse Dynamics Using Sparse Online\\Gaussian Process with Forgetting Mechanism
}

\author{Wei Li$^{2}$, Zhiwen Li$^{1}$, Yiqi Liu$^3$, and Yongping Pan$^{*1}$ 
\thanks{*Corresponding author. This work was supported in part by the Guangdong Pearl River Talent Program of China under Grant No. 2019QN01X154}
\thanks{$^{1}$Z. Li and Y. Pan are with the School of Computer Science and Engineering, Sun Yat-sen University, Guangzhou 510006, China {\tt\small lizhw63@\-mail2.sysu.edu.cn; panyongp@mail.sysu.edu.cn}}
\thanks{$^{2}$W. Li is with the School of Software Engineering, Sun Yat-sen University, Guangzhou 510006, China {\tt\small liwei363@mail2.sysu.edu.cn}}
\thanks{$^{3}$Y. Liu is with the School of Automation Science and Engineering, South China University of Technology, Guangzhou 510640, China  {\tt\small aulyq@sc\-ut.edu.cn}}
}

\begin{document}

\maketitle

\begin{abstract}

Online Gaussian processes (GPs), typically used for learning models from time-series data, are more flexible and robust than offline GPs. Both local and sparse approximations of GPs can efficiently learn complex models online. Yet, these approaches assume that all signals are relatively accurate and that all data are available for learning without misleading data. Besides, the online learning capacity of GPs is limited for high-dimension problems and long-term tasks in practice.
This paper proposes a sparse online GP (SOGP) with a forgetting mechanism to forget distant model information at a specific rate. The proposed approach combines two general data deletion schemes for the basis vector set of SOGP: The position information-based scheme and the oldest points-based scheme.
We apply our approach to learn the inverse dynamics of a collaborative robot with 7 degrees of freedom under a two-segment trajectory tracking problem with task switching. Both simulations and experiments have shown that the proposed approach achieves better tracking accuracy and predictive smoothness compared with the two general data deletion schemes.

\end{abstract}

\section{Introduction}
Accurate models of robot dynamics are significant for both exact tracking and compliant control, yet mechanism modeling is costly for many real-world robots, and reliable models are still difficult to obtain in many scenarios.
Gaussian process (GP) is a data-driven, interpretable, and nonparametric model that requires fewer training points than neural networks in modeling and provides uncertainty estimates \cite{williams2006gaussian}.
Considerable works have been done on applying GPs to learn inverse robot dynamics for control purposes, e.g., see \cite{nguyen2010using, um2014independent, alberto2014computed, beckers2019stable, chen2019gaussian, rezaei2019cascaded, lima2020sliding}. However, the applied GPs and many other GPs \cite{quinonero2005unifying} are trained offline.
Offline GPs require lots of training data to establish global models and are helpless for model variations or environmental changes \cite{meier2016drifting}.
Online GPs are capable of learning robot models fast and online from scratch and hence have better adaptability and robustness than offline GPs \cite{liu2020gaussian, prando2016online}. The major challenge limiting online GPs for real-time applications comes from their high computational complexity in solving matrix inversion for model update and prediction.

One major direction to tackle the computational complexity problem is to employ multiple local GPs. In the seminal work \cite{nguyen2008local} of this direction, a number of local GPs are generated online, and a predicted output is calculated by weighting the distance from local GP centers.
Some approaches consider tree structures for local GPs \cite{park2013learning, lederer2021gaussian}. The approach of \cite{park2013learning} uses a multi-layer tree structure to store local GPs and to make prediction by the closest model to the incoming point. The approach of \cite{lederer2021gaussian} stores and activates GPs using random trees and achieves a logarithmic complexity to the number of training points for model update and prediction.
The approach of \cite{meier2016drifting} performs multiple local GPs in parallel and proposed two possible ways of combining them for prediction, where each GP has a different neighbor size.
Yet, most local GP models sacrifice global information to some extent in the pursuit of running speed \cite{gijsberts2013real}.

Applying sparse approximations of GPs, which pays special attention to global features, is another major direction to tackle the computational complexity problem.
This direction reduces computational cost by constructing a small data set \cite{csato2002sparse, de2012online} or approximating some parts of the computational process \cite{bui2017streaming, gijsberts2013real, romeres2016online, romeres2019derivative}.
In \cite{csato2002sparse}, an incremental learning model termed sparse online GP (SOGP) was proposed to make real-time prediction by maintaining a small subset of online data.
Cruz et al. \cite{de2012online} considered the update of hyperparameters for SOGP. Bui et al. \cite{bui2017streaming} applied $\alpha$-divergence instead of Kullback-Leibler divergence to SOGP regression and provided algorithms to optimize hyperparameters and pseudo-input locations.
The inverse of the kernel was bypassed through spectral decomposition in \cite{gijsberts2013real, romeres2016online, romeres2019derivative}.
A corplane learning scheme of SOGP was proposed in \cite{deng2022data} to reduce the kernel function computation and was applied to impedance learning for robot interaction control in \cite{deng2021sparse}.
In \cite{wilcox2020solar},  sparse, local, and streaming techniques were combined to build sparse online locally adaptive regression using GPs, which processes data on localized sparse GPs.

It is still tricky for the abovementioned online GPs to meet the requirement of real-time computation in robot control applications.
Besides, most online GPs do not consider the timeliness of their inputs. On the one hand, local GPs encounter difficulties in excluding time-varying useless data points. On the other hand, sparse approximations of GPs relying on divergence-like or norm schemes tend to retain informative points if the task switches, potentially leading to poor local performance.
During the actual running of real-world robots, task switching and unexpected disturbances frequently happen. GPs may learn some useless data points in this scenario, which is harmful for learning robot dynamics.
A reasonable forgetting scheme is worth introducing into learning models \cite{van2012kernel}, where data points that are informative but useless can be forgotten when the task is switched, bringing better prediction and control performance.

This paper introduces a SOGP with a forgetting mechanism (SOGP-FM) to trade off position information and time information, pursuing short-term performance while maintaining some long-term features.
To solve the real-time learning problem of GPs, we decrease the updating frequency and focus more on recent online data to reduce computational cost while maintaining prediction accuracy.
The proposed approach is applied to learn the inverse dynamics of a collaborative robot with 7 degrees of freedom (DoFs) called Franka Emika Panda under a two-segment trajectory tracking problem. Simulations and experiments are conducted to compare prediction accuracy and smoothness, and tracking accuracy of various GP control schemes, including the position information-based scheme (PIS), the oldest points-based scheme (OPS), and the proposed forgetting scheme (FS).


Throughout this article, $\mathbb{R}$, $\mathbb{R}^+$, $\mathbb{R}^n$ and $\mathbb{R}^{m\times n}$ denote the spaces of real numbers, positive real numbers, real $n$-vectors and real $m \times n$-matrices, respectively,
$L_\infty$ denotes the space of bounded signals,
$\mathcal{N}(\bm{c}, K)$ denotes a multivariate Gaussian distribution with a mean vector $\bm{c} \in \mathbb{R}^d$ and a covariance matrix $K \in \mathbb{R}^{d \times d}$,
$\mathcal{GP}(\bar{f}(\bm{x}), k(\bm{x}, \bm{x}'))$ is a GP with a mean function $\bar{f}: \mathbb R^n \mapsto \mathbb R$ and a kernel function $k: \mathbb R^n \times \mathbb R^n \mapsto \mathbb R$,
$\Vert \cdot \Vert^2_{\rm{RKHS}}$ is the reproducing kernel Hilbert space (RKHS) norm,
$T_{t+1}(\bm{x})$ expands a $t$-vector to a $t$+1-vector by appending zeros at the end of $\bm{x}$,
$U_{t+1}(A)$ expands a ($t \times t$)-matrix to a ($t$+1)$\times$($t$+1)-matrix by appending zeros at the last row and column of $A$,
mod$(n, m)$ denotes $n$ modulo $m$,
$\bm{x}(i)$ denotes the $i$th element of $\bm{x}$,
$A(i,j)$ denotes the element at the $i$th row and $j$th column of $A$,
$P(z)$ and $\left< P(z) \right>$ denotes the probability distribution and its expectation of a random variable $z$, respectively,
and diag($x_1$, $x_2$, $\cdots$, $x_n$) denotes a diagonal matrix with diagonal elements $x_1$ to $x_n$,
where $\bm{x} \in \mathbb R^n$, $A \in \mathbb{R}^{m\times n}$, $x_i \in \mathbb{R}$, $i =$ 1 to $n$, and $n$, $m$ and $t$ (epoch) are positive integers. Note that $t$ also denotes the continuous time for the continuous-time case.

\section{Problem Formulation}\label{problem formulation}

The dynamic model of robotic manipulators with $n$-DoFs can be represented as follows \cite{spong2008robot}:
\begin{equation}\label{eq:dynamic}
\begin{matrix}
    \underbrace{M(\bm{q})\ddot{\bm{q}} + C(\bm{q},\dot{\bm{q}})\dot{\bm{q}} + \bm{g}(\bm{q}) + \bm{\epsilon}(\bm{q}, \dot{\bm{q}}, \ddot{\bm{q}})} = \bm \tau \\
	 \bm{f}(\bm{q}, \dot{\bm{q}}, \ddot{\bm{q}})\;\;\;\;\;\;\;\;\;\;
\end{matrix}
\end{equation}
where $\bm q(t) = [q_1(t), q_2(t), \dots, q_n(t)]^T \in \mathbb R^n$ is a joint angular position, $M(\bm{q}) \in \mathbb{R}^{n \times n}$ is an inertia matrix, $C(\bm{q},\dot{\bm{q}}) \in \mathbb{R}^{n \times n}$ is a centripetal-Coriolis matrix, $\bm{g}(\bm{q}) \in \mathbb{R}^{n}$ is a gravitational torque, $\bm{\epsilon}(\bm{q}, \dot{\bm{q}}, \ddot{\bm{q}}) \in \mathbb{R}^{n}$ is the unmodeled dynamics, and $\bm{\tau}(t) \in \mathbb{R}^n$ is a control torque.
The whole dynamic model can be considered as a nonparametric model $\bm \tau$ $=$ $\bm{f}(\bm{q}, \dot{\bm{q}}, \ddot{\bm{q}})$. Then, GP can be applied to learn this inverse dynamic model to perform model-based control.
Let $\bm q_d(t)$ $:=$ $[q_{d1}(t)$, $q_{d2}(t)$, $\dots$, $q_{dn}(t)]^T \in \mathbb R^n$ denote a desired output satisfying $\bm q_d(t)$, $\dot{\bm q}_d(t)$, $\ddot{\bm q}_d(t)$ $\in L_\infty$.
The control law $\bm{\tau} \in \mathbb{R}^n$ consisting of a feedforward term $\bm{\tau}_{\mathrm{ff}} \in \mathbb{R}^n$ and a feedback term $\bm{\tau}_{\mathrm{fb}} \in \mathbb{R}^n$ can be applied as follows:
\begin{equation}\label{eq:control_torque}
\begin{split}
    \bm{\tau} = \underbrace{\bm{f}_{\mathrm{GP}}(\bm{q}_d, \dot{\bm{q}}_d, \ddot{\bm{q}}_d)}_{\bm{\tau}_{\mathrm{ff}}} + \underbrace{K_{p}\bm{e} + K_{d}\dot{\bm{e}}}_{\bm{\tau}_{\mathrm{fb}}}
\end{split}
\end{equation}
in which $\bm{f}_{\mathrm{GP}}(\bm{q}_d, \dot{\bm{q}}_d, \ddot{\bm{q}}_d):=[f_{\mathrm{GP}1}(\bm{q}_d, \dot{\bm{q}}_d, \ddot{\bm{q}}_d), f_{\mathrm{GP}2}(\bm{q}_d, \dot{\bm{q}}_d,$ $ \ddot{\bm{q}}_d)$, $\dots$, $f_{\mathrm{GP}n}(\bm{q}_d, \dot{\bm{q}}_d, \ddot{\bm{q}}_d)]^T \in \mathbb R^n$ consists of $n$ independent GPs $f_{\mathrm{GP}i}: \mathbb R^{3n} \mapsto \mathbb R$ with $i =$ 1 to $n$, $\bm{e}(t) := \bm q_d(t) - \bm q(t)$ is a position tracking error, and $K_p, K_d \in \mathbb{R}^{n \times n}$ are positive-definite diagonal matrices of control gains.
With accurate model predictions, only a low gain of the feedback term $\bm{\tau}_{\mathrm{fb}}$ is required to ensure closed-loop stability and exact tracking.

\section{Sparse Online Gaussian Process}\label{sogp}

The SOGP presented in \cite{csato2002sparse} is summarized in this section.
Consider a data set $(X, \bm{y})$ generated by $y_i=f(\bm{x}_i)+\varepsilon_n$ with $f:\mathbb R^d \mapsto \mathbb R$, where
$X := [\bm{x}_1, \bm{x}_2, \cdots, \bm{x}_N]^T \in \mathbb{R}^{N \times d}$,
$\bm y$ $:= [y_1, y_2, \cdots$, $y_N]^T$ $\in$ $\mathbb{R}^N$,
$N$ is the number of data points,
$d$ is the number of input dimension,
and $\varepsilon_n \sim \mathcal{N}(0, \sigma_n^2)$ with $\sigma_n \in \mathbb{R}^+$.
Let $f \sim \mathcal{GP}(0, k(\bm{x}, {\bm{x}'}))$. Then, one can formulate $\bm{y} \sim \mathcal{N}(\bm{0}, K_{XX}+\sigma_n^2 I)$, where $K_{XX} := K(X, X) \in \mathbb{R}^{N \times N}$ denotes a covariance matrix given by
\begin{equation}
K(X, X) := \begin{bmatrix}
k(\bm x_{1}, \bm x_{1}) & \cdots & k(\bm x_{1}, \bm x_{N})\\
\vdots & \ddots & \vdots \\
k(\bm x_{N}, \bm x_{1}) & \cdots & k(\bm x_{N}, \bm x_{N})
\end{bmatrix}
\end{equation}
and $k(\cdot,\cdot)$ is frequently chosen as a Gaussian kernel
\begin{equation}\label{eq:kernel}
    k(\bm{x}, \bm{x}') = \sigma_s^2 \exp \left(-(\bm{x}-\bm{x}')^T \Lambda (\bm{x}-\bm{x}')/2 \right)
\end{equation}
where $\sigma_s^2 \in \mathbb{R}^+$ is a signal variance, and $\Lambda :=$ diag($l_1,\dots,l_d$) $ \in \mathbb{R}^{d \times d}$ with $l_i \in \mathbb R^+$ ($i =$ 1, 2, $\cdots$, $d$) is a length-scale matrix of the input vector \cite{williams2006gaussian}. For a new input $\bm{x}_* \in \mathbb{R}^d$, there exists a joint Gauassian distribution
\begin{equation}\label{eq:joint dist}
    \begin{bmatrix}
    \bm{y} \\ f(\bm x_*)
    \end{bmatrix}
    \sim \mathcal{N}
    \left(\bm 0,
    \begin{bmatrix}
    K_{XX} + \sigma_n^2 I & K_{*X}^{T} \\
    K_{*X} & K_{**}
    \end{bmatrix}
    \right)
\end{equation}
to make a prediction $f(\bm x_*)$, in which $K_{*X} := K(\bm{x}_*,X) \in \mathbb{R}^{1 \times N}$, and $K_{**} := K(\bm{x}_*, \bm{x}_*) \in \mathbb{R}$.
From (\ref{eq:joint dist}), the conditional distribution can be deduced as follows:
\begin{equation}\label{eq:cond dist}
\left\{
\begin{aligned}
    & P(f(\bm x_*)|\bm{y})=\mathcal{N}(\mu_*,\sigma_*^2) \\
    & \mu_* = K_{*X} K_{XX}^{-1} \bm{y} \\
    & \sigma_*^2 = K_{**} - K_{*X} K_{XX}^{-1} K_{*X}^T
\end{aligned}
\right.
\end{equation}
with $\mu_* \in \mathbb{R}$ and $\sigma_* \in \mathbb{R}^+$.
The optimizable hyperparameters include $\sigma_n$, $\sigma_s$ and $\Lambda$, generally obtainable by iteratively optimizing the marginal likelihood. 

As the size of the data set increases, the computation cost of the matrix inversion $K_{XX}^{-1}$ in (\ref{eq:cond dist}) becomes unbearable in practice. SOGP can learn online a basis vector set (denoted by $\mathcal{BV}$) with the size $m$ ($m \ll N$) for prediction via some adding and deleting schemes \cite{csato2002sparse}.
An iterative form of GP is given by
\begin{equation}
\left\{
\begin{aligned}
& \overline{f}_{t+1}(\bm{x}) = \overline{f}_t(\bm{x}) + q_{t+1}k_t(\bm{x},\bm{x}_{t+1}) \\
  & k_{t+1}(\bm{x},\bm{x}') = k_t(\bm{x},\bm{x}')\\
& + r_{t+1}k_t(\bm{x},\bm{x}_{t+1})k_t(\bm{x}_{t+1},\bm{x}') \\
& q_{t+1} = \frac{\partial}{\partial \overline{f}_t(\bm{x}_{t+1})} \ln \left< P(y_{t+1}|f_{t+1}) \right>_t \\
& r_{t+1} = \frac{\partial^2}{\partial \overline{f}^2_t(\bm{x}_{t+1})} \ln \left< P(y_{t+1}|f_{t+1}) \right>_t
\end{aligned}
\right.
\end{equation}
with $q_{t+1}, r_{t+1} \in\mathbb{R}$, where the subscript $t$ denotes the number of iteration (i.e., epoch) in the discrete-time case.
The above recursion can be unfolded as follows:
\begin{subequations}\label{eq:recursion}
\begin{align}
& \overline{f}_t(\bm{x}) = \bm{\alpha}_t^T \bm{k}_{X_t}(\bm{x}) \label{eq:recursion_mean} \\
& k_t(\bm{x},\bm{x}') = k(\bm{x},\bm{x}') + \bm{k}_{X_t}^T(\bm{x}) C_t \bm{k}_{X_t}(\bm{x}') \label{eq:recursion_var}
\end{align}
\end{subequations}
with $\bm{k}_{X_t}(\bm{x})$ $:=$ $[k(\bm{x}_1,\bm{x})$, $k(\bm{x}_2,\bm{x})$, $\cdots$, $k(\bm{x}_t,\bm{x})]^T \in \mathbb{R}^t$ and $X_t := [\bm{x}_1, \bm{x}_2, \cdots, \bm{x}_t]^T \in \mathbb{R}^{t \times d}$, in which $\bm{\alpha}_t \in \mathbb{R}^t$ and $C_t$ $\in$ $\mathbb{R}^{t \times t}$ are iteratively updated by
\begin{equation}\label{eq:basic_update}
\left\{
\begin{aligned}
\bm{\alpha}_{t+1} &= T_{t+1}(\bm{\alpha}_t) + q_{t+1}\bm{s}_{t+1} \\
C_{t+1} &= U_{t+1}(C_t) + r_{t+1}\bm{s}_{t+1}\bm{s}_{t+1}^T \\
\bm{s}_{t+1} &= T_{t+1}(C_t \bm{k}_{X_t}(\bm{x}_{t+1})) + \bm{e}_{t+1}
\end{aligned}
\right.
\end{equation}
where $\bm{e}_{t+1}$ is the ($t+1$)th unit vector, and $\bm{s}_{t+1} \in \mathbb R^t$ is an auxiliary variable for clarity.
The increase of data points makes it impractical to use (\ref{eq:basic_update}) directly.
For $\bm{x}_{t+1}$, if $k(\bm{x}_{t+1}, \bm{x})$ can be described as follows:
\begin{equation}
k(\bm{x}_{t+1}, \bm{x}) = \hat{\bm{e}}_{t+1}\bm{k}_{X_t}(\bm{x})
\end{equation}
with $\hat{\bm{e}}_{t+1} \in \mathbb{R}^t$ a coefficient vector to be determined later, then $\bm{s}_{t+1}$ can be replaced by
\begin{equation}
\hat{\bm{s}}_{t+1} = C_t \bm{k}_{X_t}(\bm{x}_{t+1}) + \hat{\bm{e}}_{t+1}
\end{equation}
for the update of $\bm{\alpha}_{t+1}$ and $C_{t+1}$ such that the increase of their sizes caused by $T_{t+1}$ and $U_{t+1}$ does not occur. In this case, the dimensions of $\bm{\alpha}_{t+1}$ and $C_{t+1}$ are still $t$ and $t \times t$, respectively.
Thus, we can introduce a error measure
\begin{equation}\label{eq:rkhs_error}
\Vert k(\bm{x}_{t+1}, \bm{x}) - \hat{\bm{e}}_{t+1}\bm{k}_{X_t}(\bm{x}) \Vert^2_{\mathrm{RKHS}}.
\end{equation}
Minimizing (\ref{eq:rkhs_error}), one can derive
\begin{subequations}\label{eq:e_t1}
\begin{align}
& \hat{\bm{e}}_{t+1} = K^{-1}_{X_t X_t} K_{X_t \bm x_{t+1}} \label{eq_13a}, \\
& \hat{k}(\bm{x},\bm{x}_{t+1}) = \hat{\bm{e}}_{t+1}\bm{k}_{X_t}(\bm{x}) \label{eq_13b}
\end{align}
\end{subequations}
where $\hat{k}(\bm{x}_{t+1}, \bm{x}) \in \mathbb R$ is the orthogonal project of $k(\bm{x}_{t+1}, \bm{x})$. Then, define a residual error of $\hat{k}(\bm{x}_{t+1}, \bm{x})$ as follows:
\begin{equation}\label{eq:gamma}
\gamma_{x_{t+1}} := K_{{x_{t+1}}{x_{t+1}}}-K_{{x_{t+1}} X_t} K_{X_t X_t}^{-1} K_{{x_{t+1}} X_t}^{T}
\end{equation}
to determine whether to add $\bm{x}_{t+1}$ into $\mathcal{BV}$ or not.
If $\gamma_{x_{t+1}}$ is over a predefined hyperparameter $\epsilon_{\mathrm{tol}} \in \mathbb{R}^+$, $\bm{x}_{t+1}$ is added to $\mathcal{BV}$.
Note that $\gamma_{x_{t+1}}$ can be considered as the ``novelty'' measure of $\bm{x}_{t+1}$ for the current model.
This scheme tends to learn data points that have not been seen before and are far from the current model.
The above SOGP is a basic implementation of each independent GP in $\bm{f}_{\mathrm{GP}}(\bm{q}_d, \dot{\bm{q}}_d, \ddot{\bm{q}}_d)$.

\section{SOGP with Forgetting Mechanism}\label{sogp-fm}

For the point deletion in the case with size($\mathcal{BV}$) $> m$, one can remove a data point according to certain position-based information metrics, such as the RKHS norm mentioned in \cite{csato2002sparse}, denoted as the PIS.
It is also intuitive to delete the oldest point in $\mathcal{BV}$ to focus more on the recent performance, denoted as the OPS.
These two point deletion schemes are based on different ideas: The PIS keeps data points likely to be spatially useful, whereas the OPS keeps data points likely to be timely useful.
However, if a data point is spatially useful but not used in any recent task, the PIS would greatly distract the GP model from its expressive power and make it less expressive to fit complex curves locally. Moreover, if the GP model is affected by impulsive disturbances, the PIS without forgetting ability (e.g., when the RKHS norm is used) would treat unexpected data points as informative points to mislead the GP model.
On the other hand, the OPS leads to a loss of long-term predictive capability. Data points that have passed not long ago would quickly be forgotten and need to be learned again, leading to an increased modeling error each time they are relearned as well as instability in some positions.

\begin{figure}[!b]
  \centering
  \includegraphics[width=3.4in]{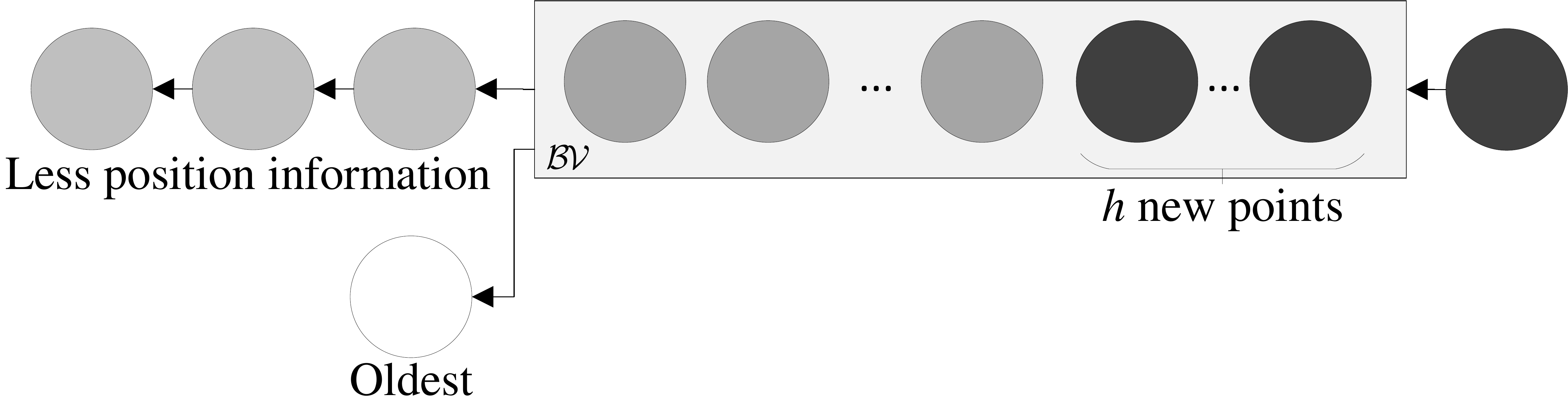}
  \caption{An online GP model that gradually forgets the oldest points at the forgetting period $h$ while retaining the more informative ones.}
  \label{fig:sogp-fm}
\end{figure}

Integrating the above two deletion schemes, we propose a SOGP-FM that adds a forgetting period $h$ while considering the spatial importance of points so that the model has short-term memory and forgets oldest points.
Let the point with Index 1 in $\mathcal{BV}$ be the earliest point added to $\mathcal{BV}$. The point with Index $i$ is selected for deletion by the following rule:
\begin{equation}\label{eq:select_index}
i=
\left\{
\begin{aligned}
&\arg\min_j \frac{\bm{\alpha}_{t+1}(j)}{K_{X_{t+1}X_{t+1}}(j, j)}, & \textrm{mod}(N, h) \neq 0 \\
&1, & \textrm{mod}(N, h) = 0
\end{aligned}
\right..
\end{equation}
When $\textrm{mod}(N, h) \neq$ 0, the point providing the least information from the position-based information metric is selected; otherwise, the earliest point added to $\mathcal{BV}$ (i.e., the oldest point) is selected. The detailed derivation can be found in \cite{csato2002sparse}. This is similar to the nature of human memory, which keeps a certain rate of forgetting to handle the short-term task and to ignore disturbances \cite{monteforte2010dynamical}.
As shown in Fig. \ref{fig:sogp-fm}, when $h$ data points are added to $\mathcal{BV}$, the oldest point or the point with less position information is deleted. Note that the oldest point is forgotten only when the GP is constantly updated.
The SOGP-FM is convenient to model time-varying dynamics and focuses more on local performance, providing better online learning for multiple tasks.
As $h\to 1$, the FS is equivalent to the OPS, and the GP keeps forgetting the oldest point in time; as $h \to \infty$, the FS is equivalent to the PIS.






\begin{figure}[!tb]
  \centering
  \includegraphics[width=3.4in]{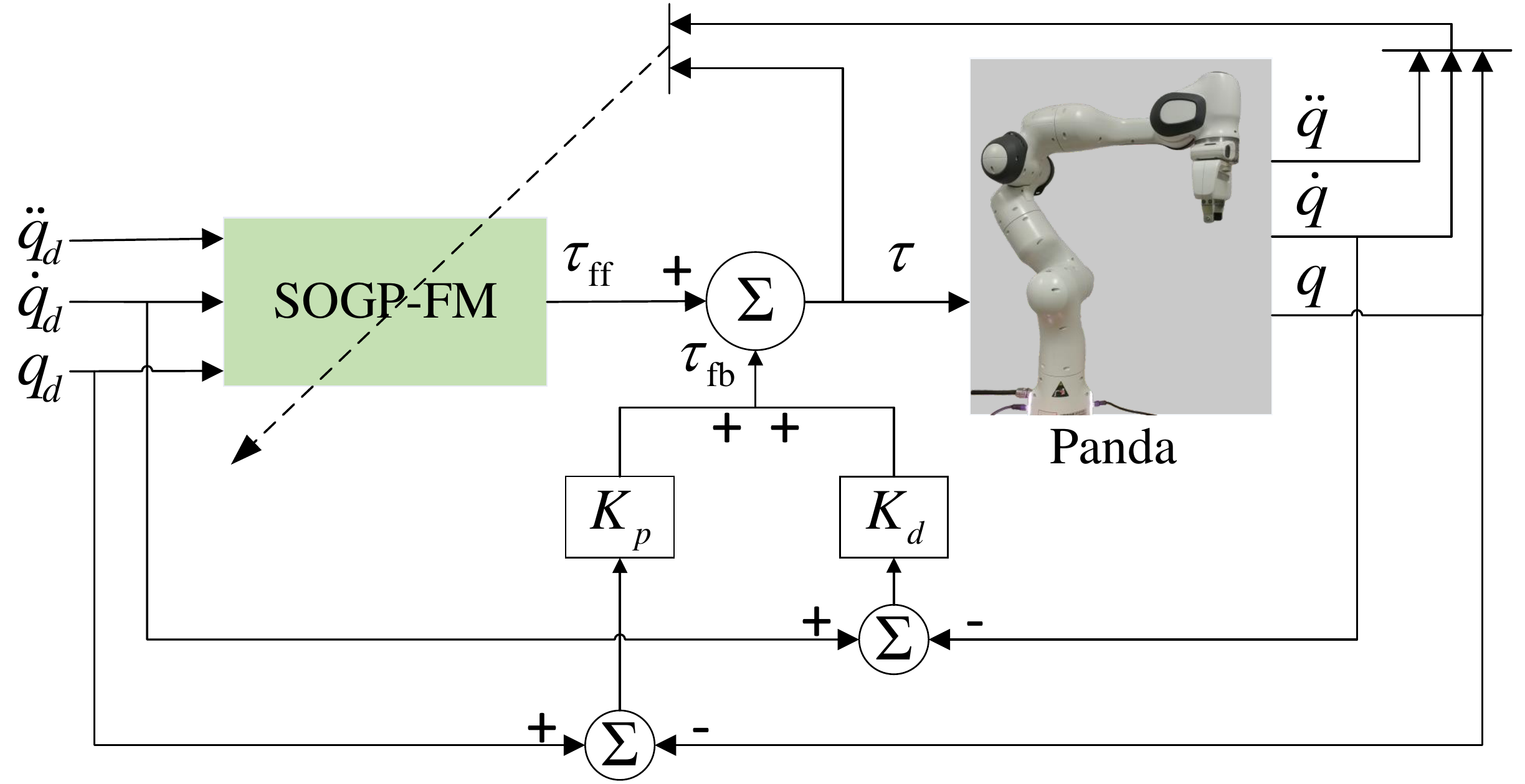}
  \caption{A block diagram of robot control based on SOGP-FM that predicts the feedforward term $\bm{\tau}_\mathrm{ff}$ and is updated online by $(\bm{q}, \dot{\bm{q}}, \ddot{\bm{q}})$ and $\bm{\tau}$.}
  \label{sogp-fm_diagram}
\end{figure}

\begin{figure}[!b]
  \centering
  \includegraphics[width=3.4in]{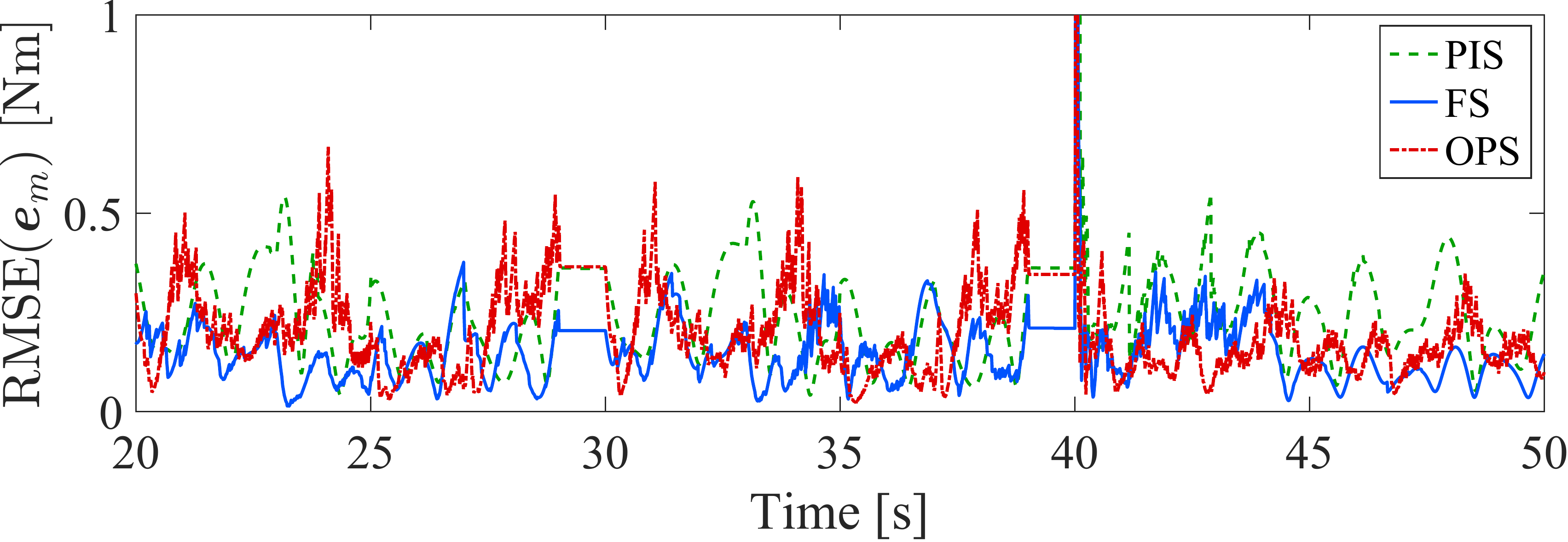}
  \caption{Modeling accuracy as RMSE($\bm{e}_{m}$) by the three SOGP schemes in simulations, where Task 2 starts at $t=$ 40 s. Note that the proposed FS has the best modeling accuracy in both tasks.}
  \label{model_err}
\end{figure}

\begin{figure}[!b]
  \centering
  \includegraphics[width=3.4in]{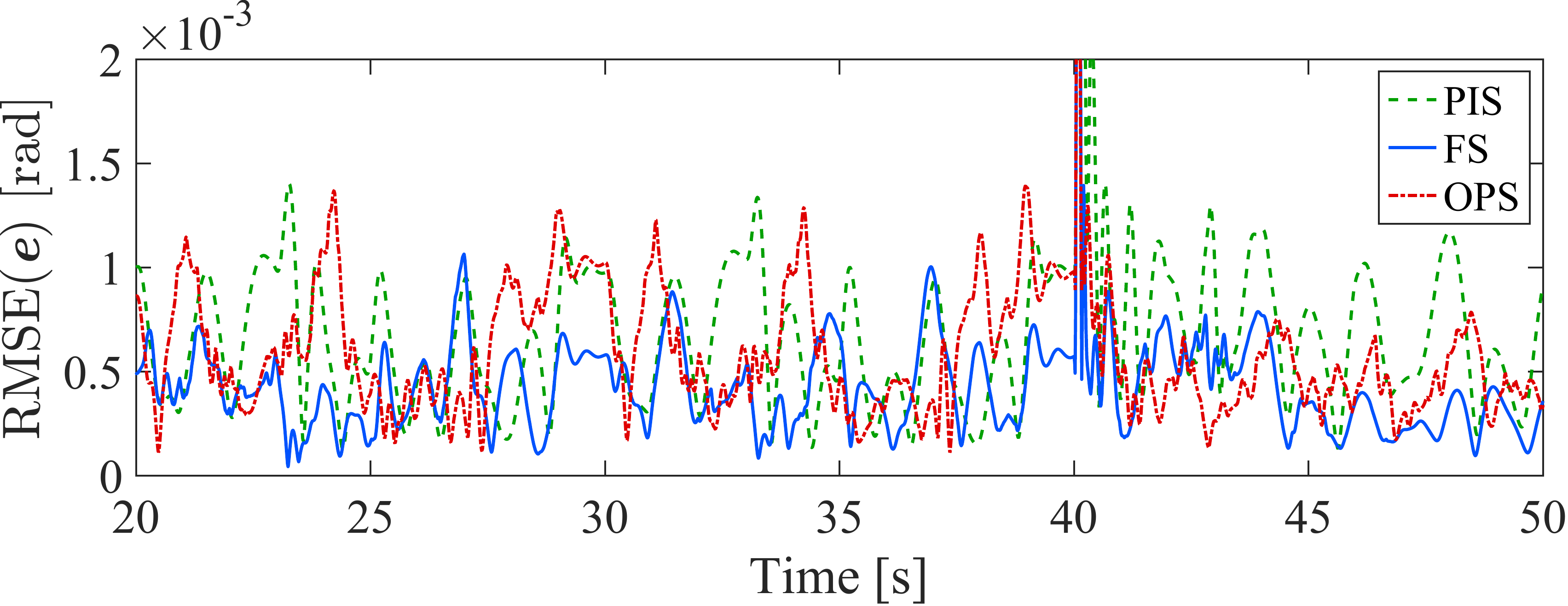}
  \caption{Position tracking accuracy as RMSE($\bm{e}$) by the three SOGP schemes in simulations, where Task 2 starts at $t=$ 40 s. Note that the proposed FS has the best tracking accuracy in both tasks.}
  \label{sim_e_norm}
\end{figure}

Generally, $h$ is set as 10$\sim$100\% of the size($\mathcal{BV}$). When the task changes, the older points need to be forgotten such that the GP can follow short-term changes.
When $h$ is set small, the GP tends to delete old points to pursuit short-term performance. However, a GP that loses long-term memory would potentially relearn previous points, leading to oscillations in fixing certain positions.
When $h$ is set large, the GP prefers long-term performance, which may result in a lack of sufficient data points to support short-term details.
Thus, $h$ needs to be adjusted to meet practical situations.
Note that $\mathcal{BV}$ will not change when the GP is predicting steadily. The SOGP-FM forgets the point only when $\mathcal{BV}$ starts changing consistently, i.e., when many new data points are added to the GP.
This approach does not increase the computational complexity for update and prediction as detailed in \cite{csato2002sparse}.


\section{Application to Collaborative Robots}\label{experiment}

A block diagram of the robot control system is shown in Fig. \ref{sogp-fm_diagram}, where GPs are employed to learn the robot inverse dynamics online and to provide a predicted feedforward torque $\bm{\tau}_{\mathrm{ff}}$ in \eqref{eq:control_torque}. To verify the generality of the proposed method, GPs do not carry any prior model information. An extended state observer (ESO) in \cite{han2009pid} is employed to obtain joint velocity $\dot{\bm{q}}$ and acceleration $\ddot{\bm{q}}$.
We perform a two-segment trajectory tracking to verify the performance of the three SOGP schemes: The PIS that removes points based on the RKHS norm, the proposed FS with $h = $ 15, and the OPS that directly removes the oldest point.

\begin{table}[!b]
\caption{Position tracking accuracy as RMSE of each joint by three SOGP schemes in simulations (0.01rad)}
\begin{center}
\begin{tabular}{|c||c|c|c|c|c|c|}
\hline
\textbf{} & \multicolumn{2}{|c|}{\textbf{PIS}} & \multicolumn{2}{|c|}{\textbf{Proposed FS}} & \multicolumn{2}{|c|}{\textbf{OPS}} \\
\cline{2-7}
\textbf{Joint} & \textbf{Task 1} & \textbf{Task 2} & \textbf{Task 1} & \textbf{Task 2} & \textbf{Task 1} & \textbf{Task 2} \\
\hline

1 & \textbf{0.023} & 0.033 & 0.033 & \textbf{0.013} & 0.039 & 0.034 \\ \hline
2 & 0.171 & 0.201 & \textbf{0.102} & \textbf{0.083} & 0.162 & 0.089 \\ \hline
3 & \textbf{0.037} & 0.042 & 0.039 & \textbf{0.029} & 0.052 & 0.044 \\ \hline
4 & 0.045 & 0.127 & \textbf{0.030} & 0.113 & 0.043 & \textbf{0.099} \\ \hline
5 & 0.015 & 0.021 & 0.014 & \textbf{0.018} & \textbf{0.013} & 0.020 \\ \hline
6 & \textbf{0.021} & \textbf{0.049} & 0.035 & 0.067 & 0.042 & 0.072 \\ \hline
7 & \textbf{0.003} & 0.004 & 0.005 & \textbf{0.004} & 0.006 & 0.006 \\ \hline
\textbf{sum} & 0.318 & 0.482 & \textbf{0.262} & \textbf{0.330} & 0.361 & 0.367 \\

\hline
\end{tabular}
\label{rmse table sim}
\end{center}
\end{table}

\subsection{Simulation Results}

This section validates the proposed approach on a simulated Panda robot. Simulations are performed through the Robotics System Toolbox in MATLAB software, in which Gaussian white noise with zero mean and covariance $10^{-14}$ is injected to the position signal $\bm q$ (consistent with the physical robot), and the ESO is applied to obtain the joint velocity $\dot{\bm{q}}$ and acceleration $\ddot{\bm{q}}$ (quite noisy).
We update GPs every 7 points and set $\epsilon_{\mathrm{tol}}=$ 0.01.
In simulations, fixed and reasonable hyperparameters are used to clearly compare the effectiveness of the three deletion schemes.
The following hyperparameters are set consistently for all GPs: $\sigma_s=1$, $\sigma_n$ $=$ 0.2, $l_{1,2,\cdots,14}$ $=$ 0.5, and $l_{15,16\cdots,21}=$ 0.2. The step size for simulations is set as 1 ms.
Moreover, with the consideration of hardware limitations, size($\mathcal{BV}$) $=$ 45 is applied here. This is because control commands can not be sent in time during experiments if size($\mathcal{BV}$) $>$ 50.


Two trajectory tracking tasks are set up here to demonstrate the fast online learning performance of the proposed SOGP-FM.
Task 1 is a periodic trajectory between four points with quintic polynomial trajectory planning during 0$\sim$40s, where the four points are (0.45, 0.25, 0.6), (0.3, $-$0.4, 0.2), (0.3, $-$0.3, 0.7) and (0.4, 0.0, 0.8) in the Cartesian coordinates (orientation is not considered). Task 2 is a periodic circular trajectory with a radius of 0.25m during 40$\sim$60s.
Set the PD control gains $K_p=$ diag(400, 400, 450, 450, 100, 100, 30) and $K_d=$ diag(10, 20, 6, 8, 4, 3, 2), which are low by considering no gravity compensation. By using only the feedback control term $\bm{\tau}_{\mathrm{fb}}$ for a simulation, the tracking error of Joint 2 reaches 0.8 rad, and the tracking errors of all other joints are around 0.01 rad.
During the first 2 s of the simulation, we need to wait for the ESO to converge to a steady state, meanwhile one can collect initial data for normalizing the output of GPs.
Then, to avoid damage to the robot caused by the sudden addition of the control torque $\bm{\tau}$, we take 0.6 s to slowly inject the predicted torque $\bm{f}_{\mathrm{GP}}$ as the feedforward term $\bm{\tau}_{\mathrm{ff}}$.

The modeling accuracy of the three SOGP schemes measured by the root-mean-square error (RMSE) is shown in Fig. \ref{model_err}, where $\bm{e}_{m} := \bm{\tau} - \bm{f}_{\mathrm{GP}}$ denotes a modeling error, and the true control torque $\bm{\tau}$ is computed by (\ref{eq:dynamic}).
One observes that in Task 1, the proposed FS has highest modeling accuracy, and the OPS suffers from local oscillations; in Task 2, the proposed FS and the OPS adapt quickly to the new trajectory and perform much better than the PIS. The poor performance of the PIS can be explained by the lack of expressiveness for complex tasks when size($\mathcal{BV}$) is small.
During the task transition, the OPS learns fast but still oscillates, while the proposed FS performs more smoothly.



\begin{figure}[!bp]
  \centering
  \includegraphics[width=3.4in]{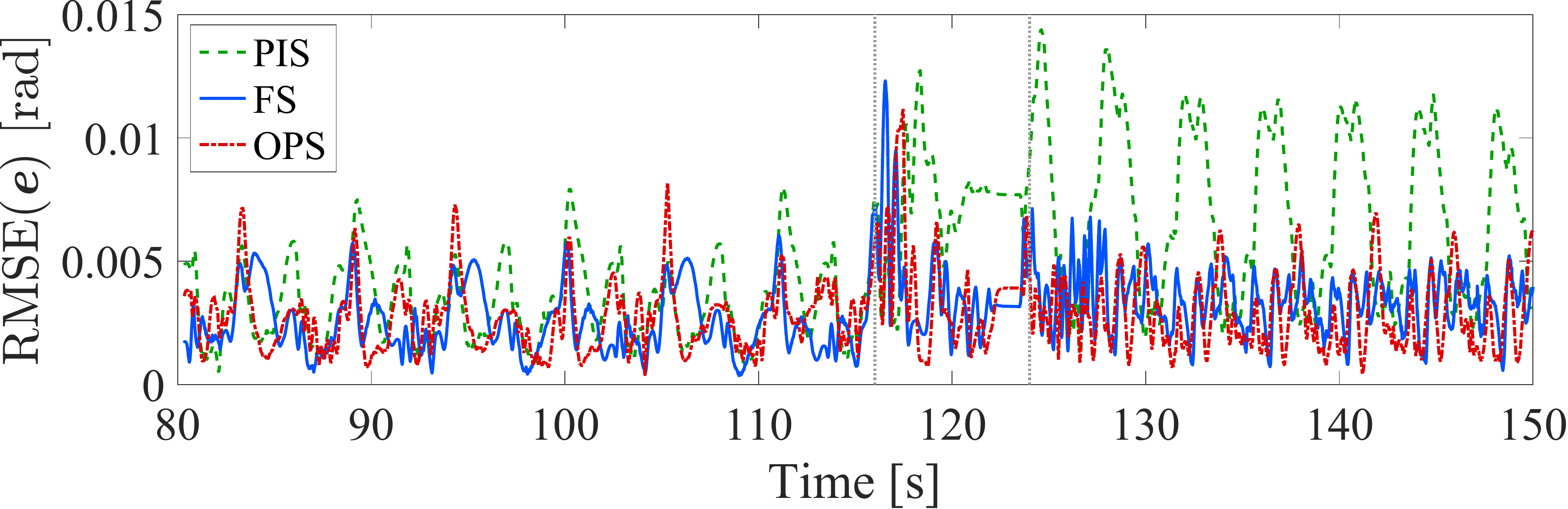}
  \caption{Position tracking accuracy as RMSE($\bm{e}$) by the three SOGP schemes in experiments, where Task 2 starts at $t=$ 124s. Note that the OPS has a larger tracking error in some areas of Task 1, and the proposed FS and the OPS outperform the PIS clearly in Task 2.}
  \label{e_norm}
\end{figure}

\begin{figure}[!b]
  \centering
  \includegraphics[width=3.4in]{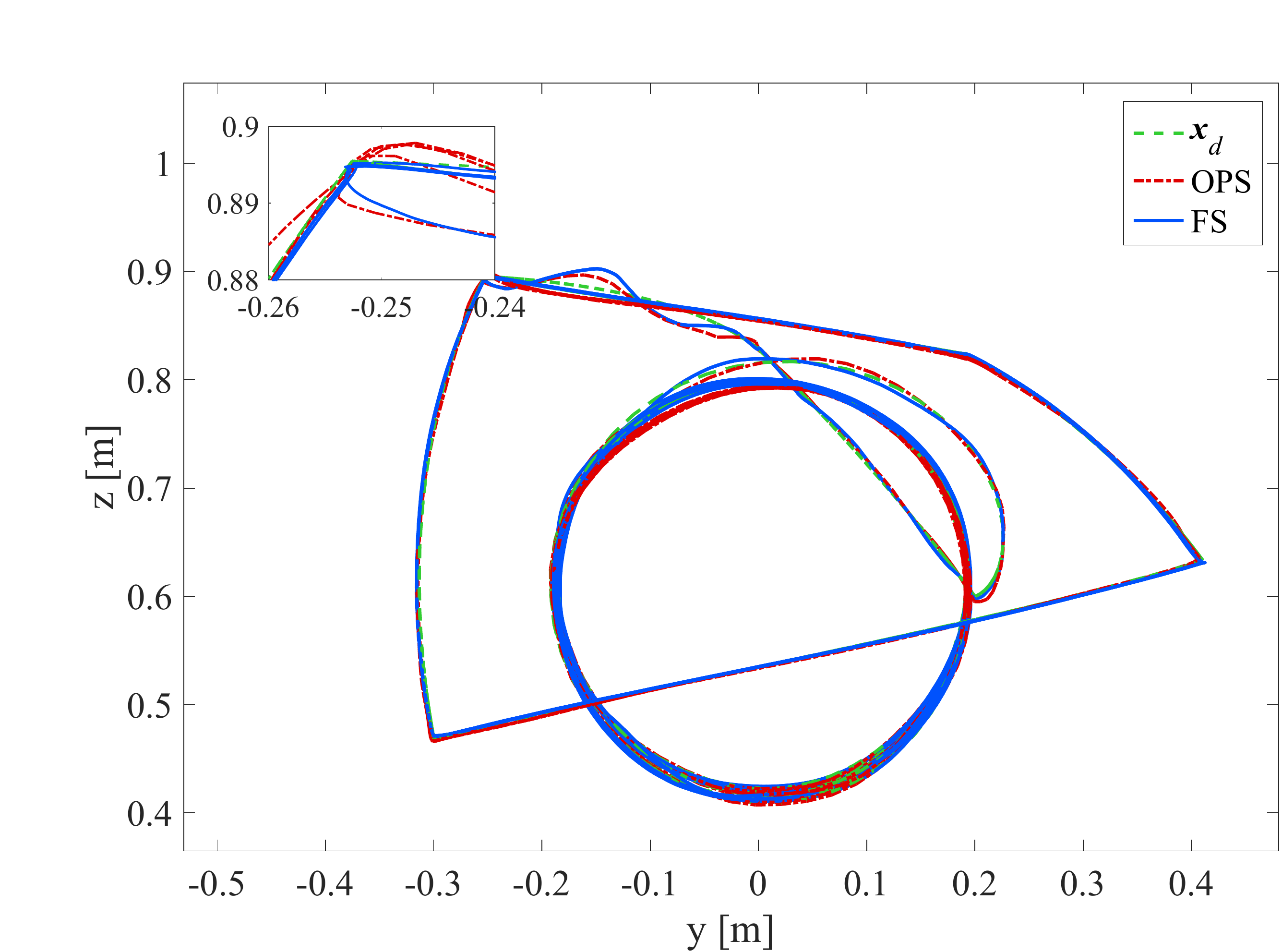}
  \caption{The Y-Z view of the end-effector in Cartesian coordinates by the OPS and the proposed FS in experiments, where $\bm{x}_d$ is the desired trajectory in the Cartesian space. Note that from the detail in the top left corner, the proposed FS performs more accurately than the OPS.}
  \label{end_effector}
\end{figure}

The control accuracy of the three SOGP schemes measured by RMSE($\bm{e}$) is shown in Fig. \ref{sim_e_norm}.
The PIS shows poor tracking accuracy in Task 1 owing to the small size$(\mathcal{BV})$ and performs much worse than the other two schemes in Task 2 due to the task switch.
The overall tracking error $\bm{e}$ of the OPS is small, but it is quite large in certain positions. This is because the OPS does not consider some essential long-term features, resulting in a transient disorder where the GP needs to relearn previously forgotten points, and further causing poor performance in some locations.
The proposed FS achieves accurate tracking in both tasks.
From TABLE \ref{rmse table sim}, the tracking error $\bm{e}$ by the PIS increases obviously after the task switch, whereas that by the proposed FS increases a little and that by the OPS increases even less.
But the OPS has a larger tracking error $\bm{e}$ in Task 1, potentially resulting from the complexity of the first trajectory that requires more long-term features.
The above results concludes that a reasonable forgetting mechanism can achieve better tracking accuracy when task changing or fast learning is required.

\begin{table}[!b]
\caption{Position tracking accuracy as RMSE of each joint by three SOGP schemes in Experiments (0.01rad)}
\begin{center}
\begin{tabular}{|c||c|c|c|c|c|c|}
\hline
\textbf{} & \multicolumn{2}{|c|}{\textbf{PIS}} & \multicolumn{2}{|c|}{\textbf{Proposed FS}} & \multicolumn{2}{|c|}{\textbf{OPS}} \\
\cline{2-7}
\textbf{Joint} & \textbf{Task 1} & \textbf{Task 2} & \textbf{Task 1} & \textbf{Task 2} & \textbf{Task 1} & \textbf{Task 2} \\
\hline

1 & 0.424 & 1.199 & \textbf{0.218} & 0.318 & 0.250 & \textbf{0.129} \\ \hline
2 & 0.563 & 1.075 & \textbf{0.555} & \textbf{0.332} & 0.381 & 0.645 \\ \hline
3 & 0.233 & 0.685 & \textbf{0.116} & 0.294 & 0.191 & \textbf{0.161} \\ \hline
4 & 0.497 & 0.569 & \textbf{0.166} & 0.359 & 0.246 & \textbf{0.295} \\ \hline
5 & 0.353 & 0.319 & \textbf{0.187} & 0.273 & 0.241 & \textbf{0.189} \\ \hline
6 & 0.404 & 0.524 & \textbf{0.290} & 0.441 & 0.359 & \textbf{0.213} \\ \hline
7 & \textbf{0.308} & 0.459 & 0.334 & 0.399 & 0.357 & \textbf{0.174} \\ \hline
\textbf{sum} & 2.786 & 4.835 & \textbf{1.870} & 2.420 & 2.029 & \textbf{1.809} \\

\hline
\end{tabular}
\label{rmse table exp}
\end{center}
\end{table}

\subsection{Experimental Results}

We build the experimental environment for the 7-DoF Panda robot by means of the Robot Operating System (ROS) and the Franka Control Interface (FCI).
Experiments consist of two tasks.
Task 1 is a periodic trajectory between four points with quintic polynomial trajectory planning during 0$\sim$116 s, where the four points are ($-$0.8, $-$0.25, 0.1, $-$1.4, 0, 1.2, 0.8), ($-$0.8, 0.25, 0.3, $-$1.8, 0.3, 1.4, 1.0), (0.8, 0.25, $-$0.1, $-$1.4, 0, 1.2, 0.8) and (0.8, $-$0.25, $-$0.3, $-$1.6, 0.3, 1.4, 1.0) in the joint space.
A transitional trajectory from Tasks 1 to 2 is set during 116$\sim$124 s.
Task 2 is a periodic circular trajectory with a radius of 0.3m during 124$\sim$150 s.
Set $\epsilon_{\mathrm{tol}} =$ 0.005, 0.004, 0.005, 0.005, 0.018, 0.02 and 0.021 for 7 GPs, respectively, and fix size$(\mathcal{BV})=$ 50. The following hyperparameters are initialized for all GPs:
$\sigma_s=1$ and $\sigma_n=0.5$, $l_{1, 2, \cdots, 14}=$ 0.5, and $l_{15, 16, \cdots, 21}=$ 0.1.
As a control command of 1 kHz is required for physical experiments, GPs are updated by polling to reduce the frequency as in the simulations. The hyperparameters are updated concurrently, where the iterative steps of the applied optimizer is limited to be 50.
Set low control PD gains $K_p =$ diag(200, 200, 200, 200, 100, 100, 50) and $K_d =$ diag(20, 20, 20, 20, 10, 10, 5), and add a gripper as the end-effector to include more unmodeled dynamics.

The control accuracy of the three SOGP schemes measured by RMSE($\bm{e}$) is shown in Fig. \ref{e_norm}, in which the results after the convergence of the hyperparameters at $t =$ 80 s are shown for clearer exhibition.
In Task 1, the PIS has acceptable performance, the OPS shows distinct fluctuations at some positions, and the proposed FS achieves the smallest tracking error $\bm{e}$.
In Task 2, the tracking error $\bm{e}$ of the PIS increases clearly, while the proposed FS and the OPS track accurately.
Note that the tracking error $\bm{e}$ of the PIS in the experiments is much larger than that in the simulations.
The proposed FS also outperforms the OPS in the Cartesian space as in Fig. \ref{end_effector}.
In TABLE \ref{rmse table exp}, the proposed FS shows the best tracking performance in Task 1 but performs worse than the OPS in some joints in Task 2. This suggests that a shorter forgetting period is required to achieve accurate tracking during the experiments. Although the OPS has smaller tracking errors in some joints, its prediction accuracy deteriorates sometimes, causing the robot to emergency stops during the experiments.


\section{Conclusions}\label{conclusion}

In this paper, we have developed a SOGP-FM which forgets distant model information
to improve the performance of SOGP and have applied it to learn the inverse dynamics of the 7-DoF Panda robot.
Simulations and experiments have validated that the proposed SOGP-FM achieves more accurate and smoother online modeling resulting in better tracking control compared with existing approaches.
However, there are still some problems in practical applications: 1) The signal-to-noise ratio for joint acceleration is small, which inevitably leads to lower modeling accuracy; 2) there are some oscillations that are expected to be alleviated during task switching. It is also interested to compare the proposed approach with neural network learning approaches \cite{RN290, RN328, RN523, RN614} for robot control in further studies.

\addtolength{\textheight}{-12cm}   

\balance
\bibliographystyle{IEEEtran}
\bibliography{root}

\end{document}